# Melanoma Classification Through Deep Ensemble Learning and Explainable AI

Wadduwage Shanika Perera[a], ABM Islam[b], Van Vung Pham[c] and Min Kyung An[d]

*Department of Computer Science, Sam Houston State University, Huntsville, Texas, U.S.A.*
*{scp030, ari014, vung.pham, an}@shsu.edu*

Keywords: Melanoma Classification, Deep Learning, Deep Ensemble Learning, Explainable AI.

Abstract: Melanoma is one of the most aggressive and deadliest skin cancers, leading to mortality if not detected and treated in the early stages. Artificial intelligence techniques have recently been developed to help dermatologists in the early detection of melanoma, and systems based on deep learning (DL) have been able to detect these lesions with high accuracy. However, the entire community must overcome the explainability limit to get the maximum benefit from DL for diagnostics in the healthcare domain. Because of the black box operation's shortcomings in DL models' decisions, there is a lack of reliability and trust in the outcomes. However, Explainable Artificial Intelligence (XAI) can solve this problem by interpreting the predictions of AI systems. This paper proposes a machine learning model using ensemble learning of three state-of-the-art deep transfer Learning networks, along with an approach to ensure the reliability of the predictions by utilizing XAI techniques to explain the basis of the predictions.

## 1 INTRODUCTION

The skin is the largest organ in the human body, and approximately a third of the total number of cancer cases are represented by skin cancers. Melanoma is the deadliest form of skin cancer, which is responsible for an overwhelming majority of skin cancer deaths. The number of melanoma deaths is expected to increase by 4.4% in 2023. Although the mortality is significant, when detected early, the 5-year survival rate for melanoma is over 99% (American Cancer Society, 2022). Currently, the most accurate way to diagnose melanoma is a biopsy. This is a penetrative surgical procedure that involves higher costs but also incorporates risks of developing various infectious diseases (Lakhtakia et al., 2009). Thus, the usual clinical practice of melanoma diagnosis is visual inspection using Dermoscopy by dermatologists or specially trained clinicians. This approach presents challenges, primarily due to its resource-intensive nature in terms of time and cost. This method's accuracy of melanoma diagnosis is approximately 80%, and the results differ from one specialist to another (Ichim et al., 2023).

Over the years, many non-invasive techniques have emerged for diagnosing skin lesions to detect melanoma. The focus has been mainly on automated, computer-based approaches, due to their efficiency and reduced susceptibility to errors. The newest trend has been the using deep learning and neural networks to detect and classify melanoma. Deep neural networks (DNNs) are increasingly prevalent in medical applications due to their capacity to address complex problems. Automated melanoma diagnosis using dermoscopy images provides a substantial potential use for deep learning techniques. However, melanoma detection is still challenging due to the various characteristics in the dermoscopy images such as low contrast, noise interference, and irregular boarders. In addition, difficulties arise from the lack of annotated data and class-imbalanced datasets. Moreover, the black-box nature of the DNN's decision-making process challenges the trustworthiness and reliability of the models. While the existing automated artificial intelligence approaches can make accurate predictions, they might lack transparency in explaining how they arrive at those conclusions.

[a] https://orcid.org/0009-0007-6123-1805
[b] https://orcid.org/0000-0003-2610-7343
[c] https://orcid.org/0000-0001-9702-8904
[d] https://orcid.org/0000-0002-4571-5048

In this paper we propose a machine learning framework to classify skin lesion images into malignant (melanoma) and benign (non-melanoma) classes, using ensemble learning of three state-of-the-art Deep Transfer Learning Models, Resnet-101, DenseNet-121, and Inception v3. Our goal is to improve the accuracy of the classification of melanoma using deep ensemble learning and to explain the predictions using explainable artificial intelligence (XAI) analysis that can aid the validation and transparency of the results.

## 2 RELATED WORKS

The work done earlier in the time has more focused on the segmentation of melanoma skin lesions, however, most of the recent work has moved on from segmentation and has focused on feature extraction and classification of melanoma. Authors in (Li & Shen, 2018) proposed two fully convolutional residual networks for simultaneous segmentation and classification. Research by (Harangi et al., 2018) implemented an ensemble network of transfer learning models AlexNet, VGG16, and GoogleNet for classification. Authors in (Bisla et al., 2019) employed a U-Net for segmentation and ResNet50 for melanoma detection and classification. They have adapted two Deep Convolutional generative adversarial networks to generate images to overcome the class imbalance issue. Authors in (Ali et al., 2019) avoided overfitting using image augmentation and they used VGG19-UNet and DeeplabV3+ for training.

Research by (Adegun & Viriri, 2020), proposed a system using a single deep convolutional neural network, based on an encoding-decoding principle to robustly extract defining features of melanoma. The encoder is responsible for learning general features and location information. The decoder learns the contour characteristics of melanoma. After extracting the features, a pixel-level classifier divided the lesions into two categories (melanoma and non-melanoma) using SegNet, U-Net, and FCN. Research by (Wei et al., 2020) developed a compact model based on two DCNNs (Deep Convolutional Neural Networks) MobileNet and DenseNet, which was proposed for melanoma diagnosis. A classification principle was introduced to improve detection accuracy. Also, a compact U-Net model based on the feature extraction module was proposed to segment the lesion area as precisely as possible.

Authors in (Xie et al., 2021) proposed a Multi-scale Convolutional Neural Network (CNN) that was implemented for melanoma classification. They have achieved significant performance by simultaneously inputting images of two scales into the network. A comparative analysis was done by (Sharma et al., 2022) where they compared the performance of transfer learning models such as VGG16, Vgg19, DenseNet-121, and ResNet-101. They used Adversarial training to generate synthetic data. Authors in (Ichim et al., 2023) used an ensemble learner of MobileNet, DenseNet-121, and DenseNet-169 for skin lesion classification. They have used weighted averaging and horizontal voting for ensemble learning. Authors in (Nandhini et al., 2023) extracted features from dermoscopy images using VGG16 and used the Random Forest algorithm to classify them.

While most of the studies focus on enhancing the accuracy of the models overlooking the aspect of assuring interpretability of their models, we focus on building a model considering the limited availability of data and amplifying model's explainability.

## 3 DATASETS

In this study dermoscopy images from the ISIC Challenge 2020 dataset and the ISIC Challenge 2019 dataset were used. The ISIC 2020 dataset includes 33,126 training images with metadata, and the ISIC 2019 dataset includes 25,331 training images with metadata. The images are labelled as malignant/ melanoma and benign/ non-melanoma (Figure 1).

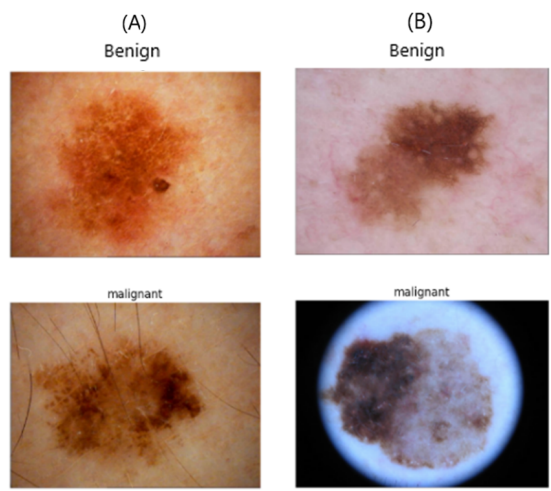

Figure 1: Examples of dermoscopy images of skin lesions from (A) ISIC 2020 dataset and (B) ISIC 2019 dataset.

ISIC 2020 dataset includes 32,542 benign images, 584 malignant images, and 27,124 images with unknown diagnoses. In this dataset, the malignant class counts for 1.8% of the whole dataset, indicating an extreme degree of class imbalance (Figure 2). ISIC 2019 dataset includes 20,809 benign images, 4,522 malignant images and, 0 images with unknown diagnoses. In this dataset, the malignant class counts for 17.8% of the whole dataset which also indicates an extreme degree of class imbalance (Figure 3).

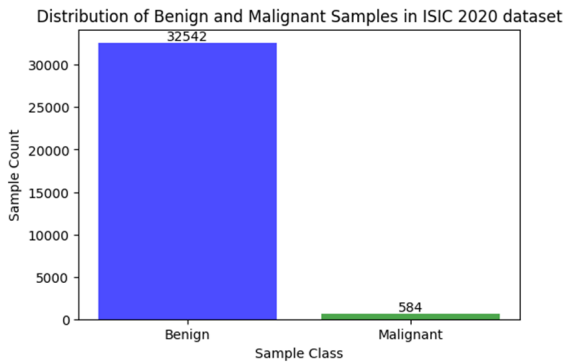

Figure 2: Class imbalance in ISIC 2020 dataset.

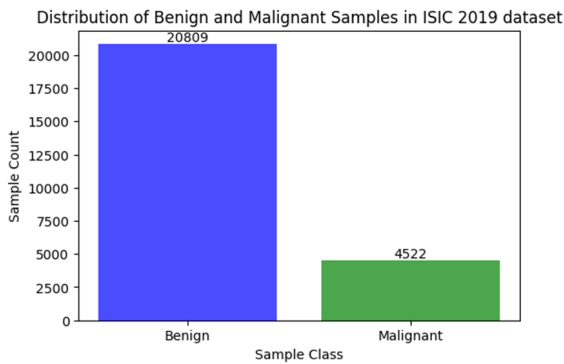

Figure 3: Class imbalance in ISIC 2019 dataset.

## 4 METHODOLOGIES

In this study, we implemented an ensembled framework of three state-of-the-art deep transfer learning neural networks, ResNet-101, Densenet-121, and InceptionV3, using a weighted average ensemble method with explainability. The methodology mainly involved five phases.

### 4.1 Data Preparation

As the first step of data preparation, the images of which the diagnosis is unknown were removed from the ISIC 2020 dataset. Then to overcome the class imbalance, ISIC 2020 and ISIC 2019 datasets were balanced by down sampling the majority class, which is the benign class, separately in two datasets and eventually combining the images of two classes (Figure 4). The final balanced dataset that was used in our work included a total of 10,212 images, where 5,106 images were benign, and 5,106 images were malignant.

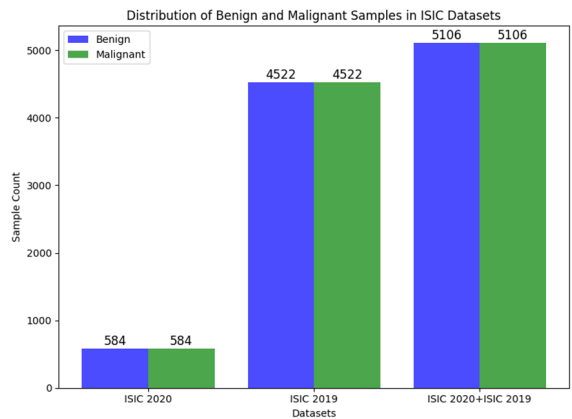

Figure 4: Balanced datasets.

### 4.2 Image Pre-Processing

Most of the images exhibited variation in quality in terms of lighting, resolution, and focus. Notably, the degree of dissimilarity between the characteristics of the skin lesion and the surrounding healthy skin was low in a considerable number of images. This fact is evident in the original images in Figure 5. Poor image quality can affect the performance of classification algorithms. Thus, the training images were enhanced using different techniques (Table 1) available in the Python Imaging Library (PIL) and OpenCV (Open-Source Computer Vision Library). The values for the factors were derived from experiments. Additionally, the images were centre cropped to mitigate position variations and normalized to eliminate redundant data and data modification errors as well as to reduce the training time.

Table 1: Image pre-processing and values used.

| Pre-processing | Factor/Value used |
|---|---|
| Colour enhancement | 1.2 |
| Sharpness enhancement | 25.0 |
| Brightness enhancement | -20 |
| Contrast enhancement | 1.5 |
| Center cropping | 0.75 of height and width |
| Normalization | Divided by 255 |

The dissimilarity or the differentiability between the characteristics of the skin lesions and the surrounding healthy skin improved after applying the above enhancements (Figure 5).

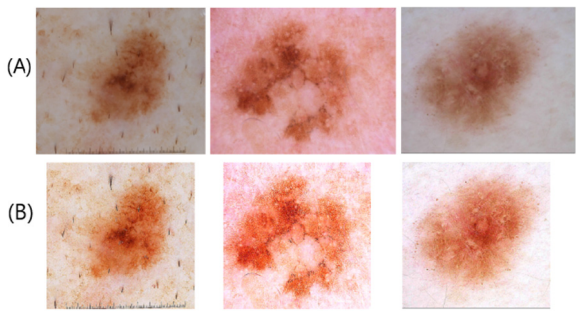

Figure 5: (A) Original images vs (B) pre-processed images.

## 4.3 Training Deep Learning Models

Five different transfer learning deep learning neural networks were trained and fine-tuned to select the best candidates as the base models for the proposed ensemble framework. They are VGG-19, ResNet-50, ResNet-101, DenseNet-121 and Inception v3.

VGG-19 architecture was proposed by (Simonyan & Zisserman, 2015), and it is a convolutional neural network (CNN) that is 19 layers deep which consists mainly of convolutional layers, followed by max-pooling layers for down sampling. In addition, the ResNet-50 and ResNet-101 proposed by (He et al., 2016) are based on residual learning frameworks. Residual blocks contain skip connections that allow the gradient to flow directly through the network, addressing the vanishing gradient problem. Besides, ResNet-50 consists of 50 convolutional layers and ResNet-101 consists of 101 of the same. Nonetheless the DenseNet-121 architecture proposed by (Huang et al., 2017) is characterized by its dense connectivity pattern with skip connections. DenseNet concatenates feature maps from different layers, leading to a more compact and computationally efficient network. The architecture has 121 trainable layers and is parameter efficient compared to other deep architectures. Inception v3, also known as GoogleNet v3, proposed by Szegedy et al., 2016) is characterized by its unique inception modules designed to capture features of multiple scales and resolutions. Inception modules combine different convolutional filter sizes within the same layer.

Transfer learning was used to train all the networks that were pre-trained using ImageNet dataset. Thus, all the images were rescaled to the size 224x224x3 as expected by the pre-trained models. All the pre-trained networks were loaded without the top output layers and custom fully connected output layers were added to make the predictions. For the custom fully connected layers, the Rectified Linear Unit (ReLU) activation function was used, and regularization was applied to the weights of the layers to eliminate overfitting. Both L1 and L2 regularization were used to further control overfitting. The networks were fine-tuned using different sets of hyperparameters. The best-performing values of the hyperparameters used to fine-tune the base learners during the training experiments are shown in Table 2.

Table 2: Hyperparameters used for training base learners.

| Hyperparameter | Value used |
|---|---|
| Batch size | 64 |
| Optimizer | Adam |
| Loss function | Categorical Cross Entropy |
| Learning rate | 0.0001 |
| Number of total epochs | 1000 |
| Early stop patience (number of epochs) | 100 |
| Activation function of the output layer | Softmax |

We applied online data augmentation to the training dataset to increase training data and overcome overfitting by exposing the model to a variety of data. We used different augmentation parameters, as shown in Table 3. The resulting augmented images after applying the augmentations on a sample image are illustrated in Figure 6.

Table 3: Data augmentation parameters and values used.

| Data Augmentation Parameter | Value used | Effect on images |
|---|---|---|
| Horizontal flip | True | Random flip through horizontal axis |
| Vertical flip | True | Random flip through vertical axis |
| Rotation | 90 | Rotation in the range of -90 to 90 while filling on the nearest pixels |
| Zoom | 0.3 | Random in and out zoom, in the proportion of 0.3 from the centre |
| Shear | 0.1 | Stretch image angle by factor of 0.1 |
| Width shift | 0.1 | Vertical random shift by 0.1 |
| Height shift | 0.1 | Horizontal random shift by 0.1 |

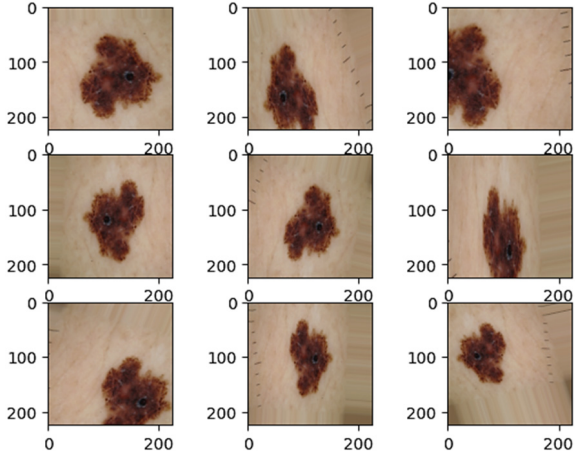

Figure 6: Different augmentations applied on a sample image.

The metrics that were calculated for the evaluation of the performance of the models are Accuracy (ACC), Precision (PRE), Recall/Sensitivity (REC), F1 score, and Area under the Receiver Operating Characteristic (ROC) curve (ROC-AUC score) as shown in Table 4. TP is the count of the true positive predictions, TN is the count of true negative predictions, FP is the count of false positive predictions and FN is the count of false negative predictions. Based on the model accuracy and the ROC-AUC score, three base learners were chosen from the experimented neural networks to build the ensemble framework.

Table 4: Evaluation Metrics.

| Metric | Formula |
|---|---|
| Accuracy | $ACC = \dfrac{TP + TN}{TP + TN + FP + FN}$ |
| Precision | $PRE = \dfrac{TP}{TP + FP}$ |
| Recall (Sensitivity) | $REC = \dfrac{TP}{TP + FN}$ |
| F1-score | $F1\ score = \dfrac{2.TP}{2.TP + FP + FN}$ |
| ROC-AUC score | $AUC - ROC\ score = \int ROC\ Curve\ (TPR\ vs.FPR)\ d(FPR)$ where, $TPR = \dfrac{TP}{TP + FN}$ $FPR = \dfrac{FP}{FP + TN}$ |

## 4.4 Ensemble Learning

Ensemble learning combines the predictions of multiple individual base learners to create a stronger, more robust model by limiting the variance and the bias issues associated with single learners with improved performance and generalization (Mienye & Sun, 2022). In our work, different fusion mechanisms of combining the predictions from the individual base learners were experimented on to find the ensemble method that best performs in classifying melanoma images. We experimented with the following four fusion methods.

### 4.4.1 Hard Majority Voting

In hard majority voting, the final class label is the class label ($c$) that is most frequently predicted by the base models (1).

$$\hat{y} = mode(c_{c1}, c_{c2}, c_{c3}) \quad (1)$$

### 4.4.2 Probability Averaging/ Soft Majority Voting

In probability averaging or soft majority voting, the maximum averaged confidence/probability is used to decide the final class prediction. The probabilities are obtained, $m$ for the malignant class (2) and $b$ for the benign class (3), and the final prediction is based on the highest probability (4).

$$m_{pred} = \frac{m_{c1} + m_{c2} + m_{c3}}{3} \quad (2)$$

$$b_{pred} = \frac{b_{c1} + b_{c2} + b_{c3}}{3} \quad (3)$$

$$\hat{y} = argmax(b_{pred}, m_{pred}) \quad (4)$$

### 4.4.3 Max Rule Ensemble Method

The classifier's prediction, which gives the maximum confidence score, is picked in the max rule ensemble method. Confidence scores are obtained, $m$ for the malignant class (5) and $b$ for the benign class (6), and the final prediction is based on the highest probability (7).

$$m_{pred} = max(m_{c1}, m_{c2}, m_{c3}) \quad (5)$$

$$b_{pred} = max(b_{c1}, b_{c2}, b_{c3}) \quad (6)$$

$$\hat{y} = argmax(b_{pred}, m_{pred}) \quad (7)$$

### 4.4.4 Weighted Probability Averaging

In weighted probability averaging, different weights are assigned to each classifier before calculating the average. Weighted average predictions are calculated, $m$ for the malignant class (8) and $b$ for the benign class (9), and the final prediction is based on the highest probability (10). The weights for three different models are $w1$, $w2$ and $w3$.

$$m_{pred} = \frac{w_1 \times m_{c1} + w_2 \times m_{c2} + w_3 \times m_{c3}}{w_1 + w_2 + w_3} \quad (8)$$

$$b_{pred} = \frac{w_1 \times b_{c1} + w_2 \times b_{c2} + w_3 \times b_{c3}}{w_1 + w_2 + w_3} \quad (9)$$

$$\hat{y} = argmax(b_{pred}, \quad m_{pred}) \quad (10)$$

The weighted average ensemble method is a highly effective fusion mechanism that is widely used. However, choosing the weights allocated to the individual base learners is a critical factor that significantly influences the ensemble's overall performance and success. Most approaches in literature set the weights experimentally or solely based on the accuracy of the base learners. (Kaleem et al., 2023). However, other evaluation measures, such as precision, recall, f1-score, and ROC-AUC score, also provide robust information for determining the importance of the base learners (Mabrouk et al., 2022). Thus, considering all the metrics in this study, we experimented with the Hyperbolic Tangent function to compute weights for our proposed ensemble framework.

For the $i^{th}$ model of the proposed ensemble framework (Figure 7), predictions were generated and compared with the true labels of the test set and the performance measures were calculated, such as precision ($pre^i$), recall ($rec^i$), f1-score ($f1^i$), and ROC-AUC score ($auc^i$). The weight of the $i^{th}$ base learner ($w^i$) was computed using the Hyperbolic Tangent function (11). The range of the hyperbolic tangent function is (0-0.762), while $m$ represents an evaluation metric, of which the values are in the range 0-1. It monotonically increases in this range; hence, if the value of a metric $m$ is high, the function acknowledges it by rewarding the weight, granting greater significance to the corresponding model; otherwise, penalizing it.

$$M^i = pre^i, rec^i, f1^i, auc^i$$

$$w^i = \sum_{m \in M^i} \tanh(m) \quad (11)$$
$$= \sum_{m \in M^i} \frac{e^m - e^{-m}}{e^m + e^{-x}}$$

The evaluation metrics mentioned in Table 4 were used to evaluate and compare the final ensemble framework.

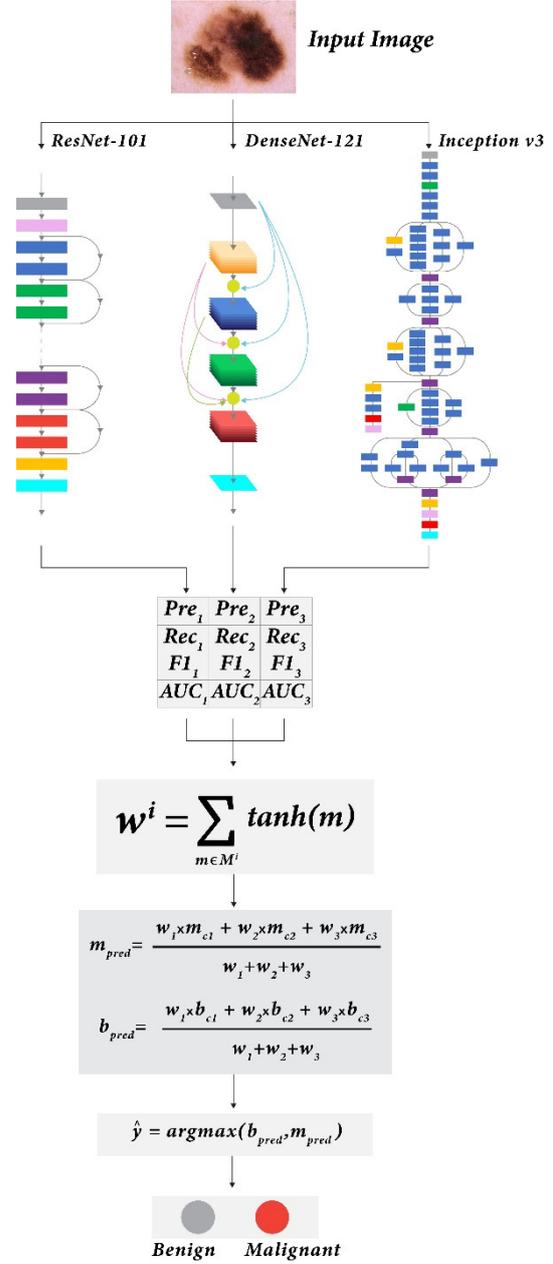

Figure 7: Proposed Ensemble learning framework.

### 4.4.5 Using SHAP for Explanation

To explore and highlight the features of the skin lesion images that contributed to the outcomes of our prediction models, we used SHapley Additive exPlanations (SHAP) analysis (Lundberg & Lee, 2017; Shakeri et al., 2021) which is a model-dependent technique. In computer vision, the SHAP values determine how much each image feature (i.e., regions, edges) contributes to the target prediction in both positive and negative directions. While most of the existing feature analysis techniques calculate the global importance of the features, the SHAP approach calculates the importance of local features for each dataset image and assigns each feature an importance value for a specific prediction. This approach can address the inconsistency problems in existing feature importance techniques and mitigate the effect of misinterpretations associated with these inconsistencies (Ian et al., 2020).

In our study, SHAP values were computed using the gradient explainer for the ensemble framework's output feature map of each base learner. We used the gradients of the base model's output feature map concerning its input features to approximate SHAP values, which provided a fair distribution of the contribution of each feature towards the prediction for a specific instance image. Then, the SHAP values were visualized on summary plots for analysis.

## 5 RESULTS AND DISCUSSION

In this section, we present the results in three sections: performance of the base classification models, performance of ensemble learning, and elaboration on the efficacy of SHAP analysis in explaining the prediction results.

### 5.1 Results of Base Learners

Table 5 displays the average values of the evaluation metrics gained by training various candidate base neural networks using a 4-fold cross-validation procedure. The networks with the highest accuracy and ROC-AUC values, Resnet-101, DenseNet-121, and Inception v3, were selected as the base models for constructing our ensemble framework.

Table 5: Performance of the candidate base models for ensemble learning.

| Model | ACC (%) | PRE (%) | REC (%) | F1-score (%) | ROC-AUC score |
|---|---|---|---|---|---|
| Vgg-19 | 73.47 | 77.68 | 66.70 | 71.77 | 0.82 |
| ResNet-50 | 77.35 | 68.87 | **94.54** | 79.69 | 0.82 |
| ResNet-101 | 80.91 | **82.53** | 78.11 | 80.26 | 0.90 |
| DenseNet-121 | **83.90** | 81.09 | 87.92 | **84.37** | **0.91** |
| Inception v3 | 81.40 | 81.63 | 80.49 | 81.06 | 0.89 |

The best-performing ResNet-101 model for the test dataset achieved a ROC-AUC score of 0.90, indicating a good separability between the two classes (Table 5). The best-performing DenseNet-121 model for the test dataset obtained the highest accuracy (83.90%) and the highest ROC-AUC score (0.91), indicating the best separability between the two classes. The best-performing Inception v3 model for the same test dataset obtained an accuracy of 81.40%, which is less than DenseNet-121 but better than ResNet-101, yet the lowest ROC-AUC score of 0.89 indicating the weakest separability between the two classes compared to the other two learners.

The confusion matrices (Figure 8) show that the ResNet-101 model demonstrated the lowest FP count, and the DenseNet-121 model resulted in the minimum FN count comparatively during the testing process. However, the Inception-v3 model maintained fewer amounts for both FP and FN counts (Figure 8).

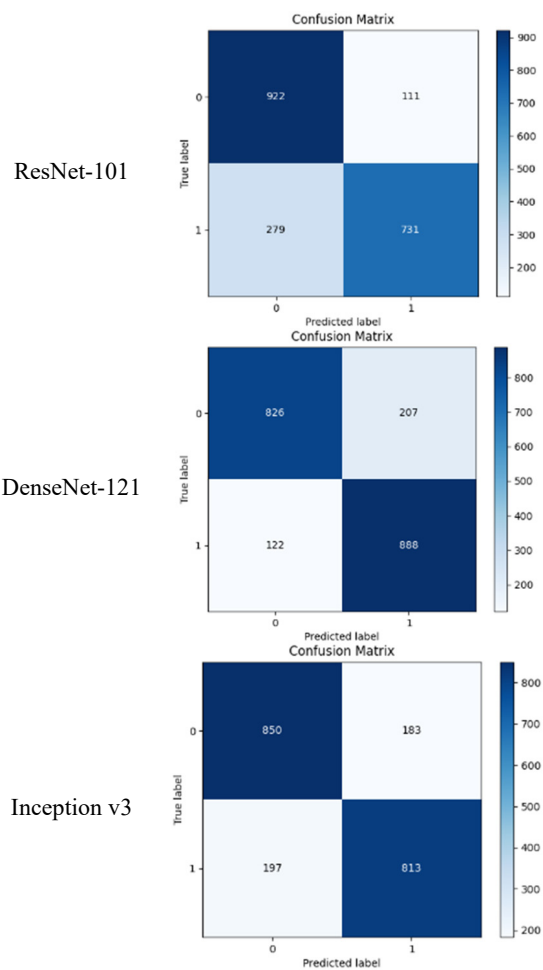

Figure 8: Confusion Matrices of the best performing models.

Table 6: Performance of different ensemble fusion methods.

| Ensemble Method | ACC (%) | PRE (%) | REC (%) | F1-score (%) | ROC-AUC score |
|---|---|---|---|---|---|
| Hard majority voting | 84.09 | 85.49 | 81.68 | 83.54 | 0.91 |
| Soft majority voting/ Probability averaging | 85.61 | **87.06** | 83.27 | 85.12 | 0.91 |
| Max rule | 85.02 | 86.67 | 82.38 | 84.47 | 0.91 |
| Weighted average with ACC as weights | 85.46 | 86.94 | 83.07 | 84.96 | 0.92 |
| Weighted average with weights computed with tanh using ACC, PRE, REC, F1 and ROC-AUC scores | 85.46 | 86.94 | 83.07 | 84.96 | 0.92 |
| Weighted average with weights computed with tanh using only PRE, REC, F1 and ROC-AUC scores | **85.80** | 86.58 | **84.36** | **85.46** | **0.93** |

Table 8: Comparison with previous work.

| Authors | Dataset | ACC (%) | PRE (%) | REC (%) | F1-score (%) | ROC-AUC score |
|---|---|---|---|---|---|---|
| (Gessert et al., 2020) | ISIC 2019 | n/a | n/a | 59.40 | n/a | 0.928 |
| (Setiawan, 2020) | ISIC 2019 | 84.76 | n/a | n/a | n/a | n/a |
| (Zhang, 2021) | ISIC 2020 | n/a | n/a | n/a | n/a | 0.917 |
| (Kaur et al., 2020) | ISIC 2016, 2017, 2020 | 82.95 | n/a | 82.99 | n/a | n/a |
| (Moazen & Jamzad, 2020) | ISIC 2019 | 84.86 | n/a | **84.85** | 46.82 | n/a |
| **Proposed method** | **ISIC 2019, 2020** | **85.80** | **86.58** | 84.36 | **85.46** | **0.93** |

## 5.2 Results of the Ensemble Learning Framework

Table 6 shows the evaluation of the ensemble learning with the five different experimental fusion mechanisms. The fusion method that performed best was the weighted averaging with the weights obtained from the hyperbolic tangent function using only precisions, recalls, f1-scores, and ROC-AUC scores of the base models, which gained the highest accuracy (85.80%) and highest ROC-AUC score (0.93). Thus, weighted averaging was chosen for the proposed ensemble framework.

The proposed ensemble learning method improved (Table 7) the overall accuracy by 1.9% and ROC-AUC score by 2% compared to the best-performing individual base learner (DenseNet-121),

Table 7: Performance of the base models and the proposed ensemble framework.

| Model | ACC (%) | PRE (%) | REC (%) | F1-score (%) | ROC-AUC score |
|---|---|---|---|---|---|
| ResNet-101 | 80.91 | 82.53 | 78.11 | 80.26 | 0.90 |
| DenseNet-121 | 83.90 | 81.09 | **87.92** | 84.37 | 0.91 |
| Inception v3 | 81.40 | 81.63 | 80.49 | 81.06 | 0.89 |
| Proposed method | **85.80** | **86.58** | 84.36 | **85.46** | **0.93** |

in classifying melanoma images into malignant and benign cases in ISIC 2020 and ISCI 2019 datasets.

Figure 9 illustrates the ROC curves of the base learners and the proposed ensemble framework based on the predictions for malignant melanoma. The proposed ensemble framework performs better than the individual networks in distinguishing malignant melanoma from benign/non-melanoma cases. The ensemble model demonstrates robustness through the smooth ROC curve (Figure 9) and its comparatively high AUC score value. It underscores the proposed model's ability to maintain predictive accuracy consistently regardless of the chosen threshold.

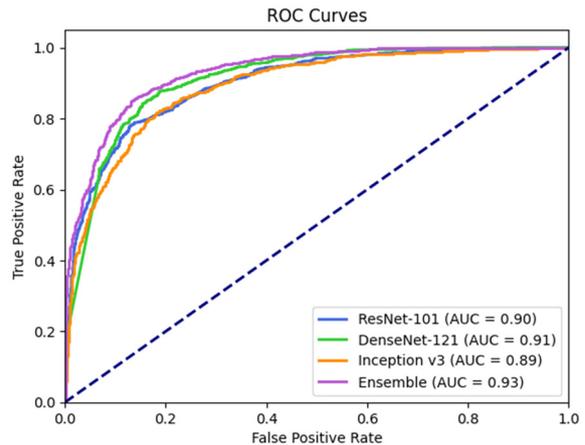

Figure 9: ROC curves of the base learners and ensemble learner.

Table 8 shows comparisons with some previous work which are outperformed by the proposed ensemble framework in classifying melanoma images.

## 5.3 Results of SHAP Analysis

In the visual representations of the SHAP analysis, pixels that correspond to the features with higher SHAP values that have a higher impact on the model's output were coloured in red, while those of the features with lower SHAP values that have a lower impact were coloured in blue. The visual representations of the SHAP analysis results for the predictions made by three base models for four sample images (A, B, C, D) are shown in Figure 10.

For all the models (Figure 10), the correct regions of interest in the sample images that can differentiate the skin lesions from the surrounding healthy skin are highlighted in red, and it shows that the relevant regions significantly contributed to the predictions of the models. This helps to increase the trustworthiness of the model. Notably, different regions of the skin lesions have impacted the outputs of the three different base models, which is evident in the results of sample C. In the resulting visual representations of the SHAP values for sample C, ResNet-101 model has given more attention to the features and edges in the bottom part of the skin lesion, while DenseNet-

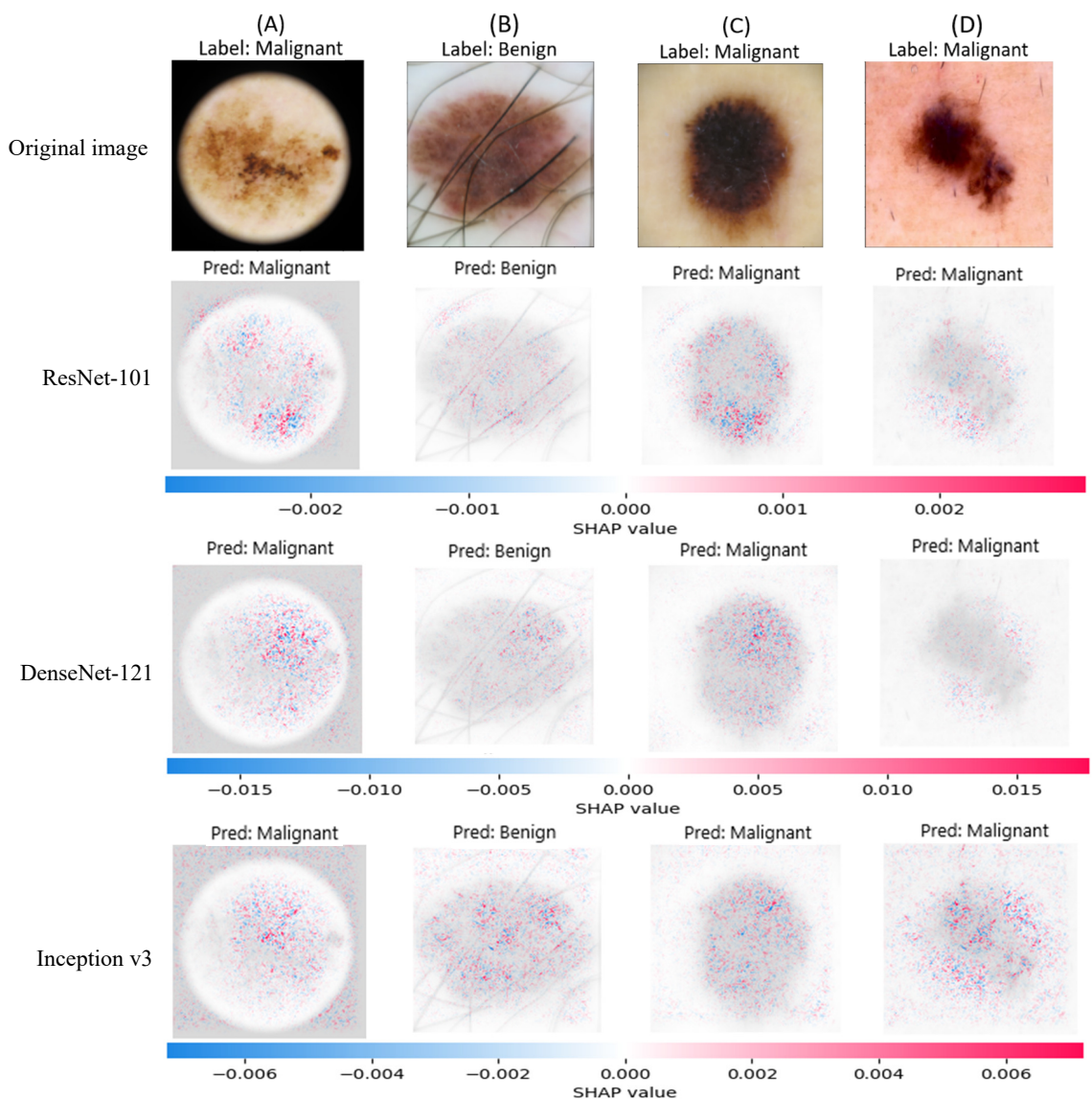

Figure 10: SHAP explanations for predictions of three base learners for samples (A), (B), (C) and (D).

121 model has considered the characteristics at the right top region of the skin lesion. For sample C, Inception v3 model has considered features from all over the skin lesion without demonstrating a specific pattern. This behaviour of Inception v3 model can be observed for all the samples A, B, C, and D.

The visual representations of the SHAP results for the DenseNet-121 model exhibit a reduced dispersion of coloured pixels and a distinctly sharper emphasis on relevant features compared to other models (Figure 10). Moreover, the highest SHAP value (0.015) of the features that contribute positively towards the DenseNet-121 model's outcome is higher than that of the other models (Figure 10). This provides an explanation for the Densent-121 model's high prediction accuracy compared to the other models.

When considering the results for sample A (Figure 10), for all three models, pixels around the edge of the circular microscopic effect are coloured in a mix of red and blue. Thus, it is evident that the microscopic effect in the image has both positively and negatively impacted the outcome of all the models. In the top left region of the SHAP results for sample B (Figure 10), the pixels around a hair that circularly curves around the lesion are coloured redder, indicating that the feature has impacted the prediction of the models. Similarly, in sample C (Figure 10) results, a pattern of red-coloured pixels can be seen forming a circular effect around the skin lesion. The models might have misinterpreted the vignetting effect in sample C as a microscopic effect around the skin lesion. Thus, it is evident that unrelated features like microscopic effects have influenced the final predictions of the models.

Moreover, as seen in the results for sample B (Figure 10), the hair atop the surface of the skin lesion is highlighted in red with higher SHAP values, thus hair has increased the probability of the class predicted for sample B. However, the hair located outside the skin lesion is not highlighted for any of the models; thus, it can be concluded that occlusion that overlaps with the region of interest had more potential to contribute to the model predictions than occlusion outside the region of interest in this study. In the results for all the samples, red and blue pixels can be seen scattered all over the surrounding healthy skin and it is apparent that the features of the surrounding healthy skin play a vital role in influencing the model output. Hence, image pre-processing that can remove significant features or occlusion in the surroundings and the region of interest can improve the performance and, most importantly, the reliability of the model.

# 6  CONCLUSIONS

Melanoma is the most lethal skin cancer type, and distinguishing between melanocytic skin lesions and melanoma in the early stages is challenging. This study provides a deep ensemble learning framework to diagnose and classify melanoma dermoscopy images with explainability. The framework's performance has been extensively evaluated using the well-recognized and publicly available datasets: ISIC 2029 and ISIC 2020. While imbalanced data and lack of labelled data are significant challenges in skin lesion classification, we have proved that with a small dataset, melanoma classification can be accomplished with competitive results by transfer learning pre-trained models. Using the weighted averaging ensemble method boosted the performances of the individual learners. SHAP explanations of the model's outcomes confirm the trustworthiness of the models by highlighting the correct regions of interest. The explanations demonstrated that the different models focus on different regions of the skin lesion to make the decision of classification, unveiling an additional advantage of using an ensemble method. Moreover, the explanations of the models prove that occlusions such as skin hair and unrelated image features such as circular microscopic effects impact the models' outcomes, indicating misinterpretation of the features of images.

Future work will focus on further improvement of the framework's performance, especially concerning the sensitivity (recall) enrichment. Sensitivity, which signifies the true positive rate, strongly influences the count of false negative cases (type II error). Therefore, enhancing sensitivity holds a significant importance within this study, given its direct relevance to medical decision-making. Additionally, occlusion removal and applying additional image pre-processing, such as lesion segmentation and colour calibration, to improve the image quality can improve the base learners' performance. The forthcoming research will also centre around validating the explanations of the model's output against the clinical features of dermoscopy images used by dermatologists to diagnose melanoma.